\documentclass[sigconf]{acmart}
\pdfoutput=1

\usepackage[utf8]{inputenc}
\usepackage[T1]{fontenc}
\usepackage{pdfrender}
\usepackage{url}
\usepackage{booktabs}
\usepackage{multirow}
\usepackage{mathtools}
\usepackage{nicefrac}
\usepackage{microtype}
\usepackage{bm}
\usepackage{color}
\usepackage{gnuplottex}
\usepackage{printlen}
\usepackage{tikz}
\usepackage{balance}

\copyrightyear{2020}
\acmYear{2020}
\setcopyright{rightsretained}
\acmConference[MileTS '20]{MileTS '20: 6th KDD Workshop on Mining and Learning from Time Series}{August 24th, 2020}{San Diego, California, USA}
\acmBooktitle{MileTS '20: 6th KDD Workshop on Mining and Learning from Time Series, August 24th, 2020, San Diego, Claifornia, USA}

\begin{document}

\title{Statistical Evaluation of Anomaly Detectors for Sequences}

\author{Erik Scharw\"achter}
\email{scharwaechter@bit.uni-bonn.de}
\orcid{0001-8555-2629}
\affiliation{%
  \institution{University of Bonn, Germany}
}

\author{Emmanuel M\"uller}
\email{mueller@bit.uni-bonn.de}
\affiliation{%
  \institution{University of Bonn, Germany}
}

\begin{abstract}
Although precision and recall are standard performance measures for anomaly detection, their
statistical properties in sequential detection settings are poorly understood.
In this work, we formalize a notion of precision and recall with temporal tolerance for
point-based anomaly detection in sequential data.
These measures are based on time-tolerant confusion matrices that may be used to
compute time-tolerant variants of many other standard measures.
However, care has to be taken to preserve interpretability.
We perform a statistical simulation study to demonstrate that precision and recall may
overestimate the performance of a detector, when computed with temporal tolerance.
To alleviate this problem, we show how to obtain null distributions for the two measures
to assess the statistical significance of reported results.
\end{abstract}


\maketitle

\section{Introduction}
\label{secIntro}

Anomaly detection in sequential data is a highly active research topic
\cite{Malhotra2015,Bridges2017,Siffer2017,Xu2018a,Ren2019,Su2019}.
Precision and recall are two measures routinely used to evaluate the performance of anomaly
detectors, both for iid data and for sequential data.
An important characteristic of sequential data is that the decisions of a detector can be
imprecise \cite{Adams2017,Adams2018a} without impairing its practical utility:
if an anomaly at time step $t$ is detected at time step $t+1$, this is still a useful result.
Recently, \citet{Tatbul2018} pointed out that the classical precision and recall measures, when
applied to sequential detection problems, may misrepresent the performance of the detector.
They introduced novel precision and recall measures for \emph{range-based} anomaly detection.
However, the problem persists even for \emph{point-based} anomalies, where the ground-truth
anomaly label is a single time step.
In this work, we study in detail time-tolerant notions of precision and recall for \emph{point-based}
anomaly detection in sequential data, with a special focus on the statistical properties of
these measures.
Our work is closely related to recent advances in the statistical association between event
series and time series \cite{Luo2014,Chi2016a,VanDortmont2019,Scharwachter2020,Scharwachter2020a},
and uses results from event coincidence analysis (ECA) \cite{Quiroga2002,Donges2016,Siegmund2016}.
Source codes are available at \url{https://github.com/diozaka/anomaly-eval}.

\subsection{Anomaly detection problem}

\begin{figure}[bt]
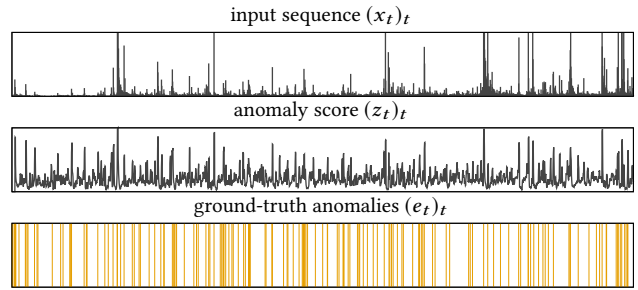

\begin{center}
\begin{gnuplot}[terminal=cairolatex,terminaloptions={size 3.33537,0.50}]
set xdata time
set timefmt "
unset tics
set format y ''
set format x ''
set lmargin screen 0.01
set rmargin screen 0.99
set tmargin screen 0.70
set bmargin screen 0.01
set yrange [0:1000]
set title '\small{input sequence $(x_t)_t$}' offset 0,-0.8
plot "data/twitter-earthquake-germany.csv" u 1:2 w i lc rgb '#444444' lw 2 notitle
\end{gnuplot}
\begin{gnuplot}[terminal=cairolatex,terminaloptions={size 3.33537,0.50}]
unset tics
set format y ''
set format x ''
set lmargin screen 0.01
set rmargin screen 0.99
set tmargin screen 0.70
set bmargin screen 0.01
set title '\small{anomaly score $(z_t)_t$}' offset 0,-0.8
plot "data/earthquakes-stalta.dat" u 0:1 w l lc rgb '#444444' lw 2 notitle
\end{gnuplot}
\begin{gnuplot}[terminal=cairolatex,terminaloptions={size 3.33537,0.50}]
set xdata time
set timefmt "
unset tics
set format y ''
set format x ''
set lmargin screen 0.01
set rmargin screen 0.99
set tmargin screen 0.70
set bmargin screen 0.01
set yrange [0:1000]
set title '\small{ground-truth anomalies $(e_t)_t$}' offset 0,-0.8
plot "data/emdat-earthquake-all.csv" u 1:($2*5000) w i lc 4 lw 1 notitle #$
\end{gnuplot}
\caption{The running example for earthquake detection.}
\label{figEarthquakes}
\end{center}
\end{figure}

We are given an input sequence $(x_t)_{t=1,...,T}$ over an arbitrary input domain.
Furthermore, we are given an anomaly scoring function $z$ to compute a numeric sequence
of anomaly scores $(z_t)_{t=1,...,T}$ from the input sequence.
If the observation at time step $t$ is likely an anomaly, the anomaly score $z_t$
should be high; if the observation at time step $t$ appears normal, $z_t$ should be low.
An anomaly is predicted at time step $t$ if the anomaly score is larger than some
predefined threshold, $z_t \ge \tau$.
The exact notion of what constitutes an anomaly is highly domain-specific and should be
reflected in the choice of the anomaly scoring function.
Anomaly detectors of this type are widely used across many disciplines.
For example, \citet{Wiedermann2016} use the clustering coefficient as an anomaly score
for dynamic networks to detect El Ni\~no events in climate data, and \citet{Earle2011} use
an energy transient score \cite{Withers1998} to detect earthquakes from
Twitter\footnote{\url{https://www.twitter.com/}} time series.

We use the problem of earthquake detection on Twitter as the running example in this work.
Figure~\ref{figEarthquakes} (top row) shows the daily volume of tweets that were posted in
Germany between 2010 and 2017 and contain the word ``earthquake,'' translated to various
languages. The plot also shows all severe earthquakes that occurred globally in the same time
period (bottom row).
We obtained the Twitter data using the ForSight platform from
Crimson Hexagon/Brandwatch\footnote{\url{https://www.brandwatch.com/}}, and the earthquake
data from the International Disaster Database EM-DAT, provided by the Centre for Research
on the Epidemiology of Disasters\footnote{\url{https://www.emdat.be/}}.
Our goal is to evaluate whether an anomaly detector on the Twitter time series has the
potential to detect earthquakes globally.
For this purpose, we stick to \citet{Earle2011} and use the energy transient score as the
anomaly score (middle row), i.e., we set $z_t = \text{STA}_t/(\text{LTA}_t+1)$, where STA
is the short-term average of the input sequence over the past 3 days, while LTA is the
long-term average over the past 14 days.
The energy transient score reacts to drastic changes in the level of the time series.

\subsection{Evaluation measures}
\begin{figure}[tb]
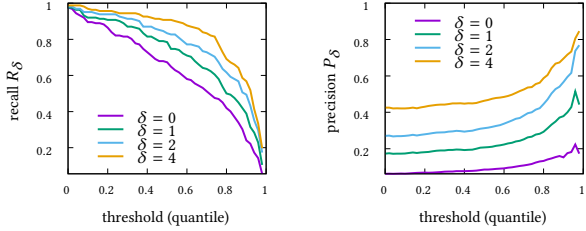

\begin{center}
\begin{gnuplot}[terminal=cairolatex,terminaloptions={size 3.33537,1.3}]
set multiplot layout 1,2
set format y '\tiny{
set format x '\tiny{
set tics scale 0.5
set ytics 0,0.2
set xtics 0,0.2 offset 0,0.5
set xlabel '\scriptsize{threshold (quantile)}' offset 0,1
set tmargin screen 0.90
set bmargin screen 0.20
set lmargin 5
set rmargin 3

set ylabel '\scriptsize{recall $R_\delta$}' offset 3,0
set key bottom left samplen 1 Left reverse spacing 0.5
plot "data/earthquakes-recall.dat" u 1:2 w l lw 3 title sprintf('\scriptsize{$\delta=0$}'), \
     "data/earthquakes-recall.dat" u 1:3 w l lw 3 title sprintf('\scriptsize{$\delta=1$}'), \
     "data/earthquakes-recall.dat" u 1:4 w l lw 3 title sprintf('\scriptsize{$\delta=2$}'), \
     "data/earthquakes-recall.dat" u 1:5 w l lw 3 title sprintf('\scriptsize{$\delta=4$}')

set ylabel '\scriptsize{precision $P_\delta$}' offset 3,0
set key top left samplen 1 Left reverse spacing 0.5
plot "data/earthquakes-precision.dat" u 1:2 w l lw 3 title sprintf('\scriptsize{$\delta=0$}'), \
     "data/earthquakes-precision.dat" u 1:3 w l lw 3 title sprintf('\scriptsize{$\delta=1$}'), \
     "data/earthquakes-precision.dat" u 1:4 w l lw 3 title sprintf('\scriptsize{$\delta=2$}'), \
     "data/earthquakes-precision.dat" u 1:5 w l lw 3 title sprintf('\scriptsize{$\delta=4$}')
\end{gnuplot}
\caption{Precision and recall by threshold and tolerance $\delta$.}
\label{figPrecRec}
\end{center}
\end{figure}

While the anomaly score encodes the \emph{feature} of interest that the anomaly
detector should react to, the detection threshold $\tau$ controls the \emph{precision and recall}
of the anomaly detector.
Let $(e_t)_{t=1,..,T}$ be a ground-truth sequence of actual anomalies, with value $e_t=1$ if
there is an actual anomaly at time step $t$, and $e_t=0$ if there is no anomaly.
In our example, actual anomalies correspond to reported earthquakes.
We define precision $P_0$ and recall $R_0$ as
\begin{gather}
\label{eqnPrecRec}
P_0 = \frac{\sum_t e_t \cdot \mathcal{I}(z_t \ge \tau)}{\sum_t \mathcal{I}(z_t \ge \tau)}
\text{ and }
R_0 = \frac{\sum_t e_t \cdot \mathcal{I}(z_t \ge \tau)}{\sum_t e_t}.
\end{gather}
The function $\mathcal{I}(c)$ evaluates to 1 if and only if the condition $c$ is true.
The numerator is the number of true positives, while the denominator is either the number of
predicted anomalies or the number of actual anomalies.
The relationship between the threshold and precision and recall for the earthquake detection
problem is visualized in Figure~\ref{figPrecRec}.
The values for $P_0$ and $R_0$ from Equation~\ref{eqnPrecRec} correspond to the lines labeled
$\delta=0$ in the plots.
The results are not particularly good: we can obtain acceptable recall values at low thresholds,
but the cost is an unacceptably low precision.
We observe that recall is a monotonically decreasing function of the threshold $\tau$:
the number of true positives in the numerator decreases with increasing $\tau$ while the
denominator stays constant.
Precision, on the other hand, is a non-monotone function of the threshold, since both the
numerator and the denominator change with $\tau$.

  \section{Time-tolerant measures}
\subsection{Precision and recall}
In sequential data, a predicted anomaly can often be considered a true positive if there is an
actual anomaly \emph{close to} the predicted time point.
The higher the temporal tolerance, the higher the number of true positives, and the higher will
be \emph{both} precision and recall.
We follow ECA \cite{Donges2016,Siegmund2016,Scharwachter2020a} and define measures
for precision $P_\delta$ and recall $R_\delta$ with temporal tolerance $\delta$ as:
\begin{gather}
\label{eqnPrecTolerant}
P_\delta = \frac{\sum_t \mathcal{I}\left(\sum_{t'=t-\delta}^{t+\delta} e_{t'}>0\right) \cdot
				\mathcal{I}(z_t \ge \tau)}{\sum_t \mathcal{I}(z_t \ge \tau)} \\
\label{eqnRecTolerant}
R_\delta = \frac{\sum_t e_t \cdot
				\mathcal{I}\left(\sum_{t'=t-\delta}^{t+\delta}\mathcal{I}(z_{t'} \ge \tau)>0\right)}{\sum_t e_t}.
\end{gather}
If $\delta=0$, the tolerant measures are equivalent to the standard measures.
If $\delta>0$, the definition of a true positive in the numerator changes.
In fact, there are now two different types of true positives: In the case of precision,
a true positive is a predicted anomaly at time step $t$ with an actual anomaly within
the range $[t-\delta,t+\delta]$. In the case of recall, a true positive is an actual anomaly
at time step $t$ with a predicted anomaly within the range $[t-\delta,t+\delta]$.

Figure~\ref{figPrecRec} shows the impact of various choices for the
temporal tolerance $\delta$ on the measured values for precision and recall.
Depending on the choice of the threshold and the temporal tolerance, the reported values
for precision and recall vary drastically.
The adoption of temporal tolerance in the evaluation is undoubtedly valid in many
applications. However, the example shows that the use of moderate temporal tolerances may already
lead to overstated performance measures that do not necessarily reflect the actual utility of
the results.
In Section~\ref{secSimulationStudy}, we perform a simulation study to further investigate this issue.

\subsection{Confusion matrices}
The extension of precision and recall to time-tolerant measures via relaxed notions
of \emph{true positives} is intuitive, but has some subtleties not discussed before.
In fact, these two measures are computed from two distinct \emph{confusion matrices}, where
temporal tolerance is allowed \emph{either} in the ground-truth time steps \emph{or} in
the predicted time steps. The general structure of a confusion matrix is:

\begin{table}[H]
\begin{center}
\begin{tabular}{r c c c}
\toprule
       & AA & AnA & \\
\midrule
PA     & $TP$ &  $FP$ & $\sum$ \\
PnA    & $F$N &  $TN$ & $\sum$ \\
\midrule
       & $\sum$ & $\sum$ & $T$\\
\bottomrule
\end{tabular}
\end{center}
\end{table}

It contains the numbers of observations that fall into the four categories
true positives ($TP$), false positives ($FP$), false negatives ($FN$)
and true negatives ($TN$), along with marginal sums.
The row and column headings define the marginal conditions: actual anomaly (AA),
actually no anomaly (AnA), predicted anomaly (PA), predicted no anomaly (PnA).
The confusion matrix partitions the observations so that every observations falls
in exactly one category.
Many performance measures can be computed from confusion matrices \cite{Powers2007},
typically by normalizing individual entries by marginal sums.
The measures are interpretable because all entries and marginals have
straightforward interpretations.

\begin{table*}
\begin{center}
\caption{Relaxed confusion matrix for sequential data, with tolerance in ground-truth}
\small
\label{tblRelConfMatG}
\begin{tabular}{r c c c}
\toprule
                & AA$\delta$ & AnA$\delta$ & \\
\midrule
PA & $\displaystyle\sum_t \mathcal{I}\left(\sum_{t'=t-\delta}^{t+\delta} e_{t'} > 0\right)
							\mathcal{I}(z_{t} \ge \tau)$
   & $\displaystyle\sum_t \left(1-\mathcal{I}\left(\sum_{t'=t-\delta}^{t+\delta} e_{t'} >0\right)\right)
							\mathcal{I}(z_{t} \ge \tau)$
   & $\displaystyle\sum_t \mathcal{I}(z_{t} \ge \tau)$ \\
PnA& $\displaystyle\sum_t \mathcal{I}\left(\sum_{t'=t-\delta}^{t+\delta} e_{t'} > 0\right)
							(1-\mathcal{I}(z_{t} \ge \tau))$
   & $\displaystyle\sum_t \left(1-\mathcal{I}\left(\sum_{t'=t-\delta}^{t+\delta} e_{t'}>0\right)\right)
							(1-\mathcal{I}(z_{t} \ge \tau))$
   & $\displaystyle\sum_t (1-\mathcal{I}(z_{t} \ge \tau))$ \\
\midrule
           & $\displaystyle\sum_t \mathcal{I}\left(\sum_{t'=t-\delta}^{t+\delta} e_{t'} > 0\right)$
           & $\displaystyle\sum_t \left(1-\mathcal{I}\left(\sum_{t'=t-\delta}^{t+\delta} e_{t'} > 0\right)\right)$
           & $T$ \\
\bottomrule
\multicolumn{4}{c}{\tiny{%
AA$\delta$: actual anomaly with tolerance $\delta$,
AnA$\delta$: actually no anomaly with tolerance $\delta$,
PA: predicted anomaly,
PnA: predicted no anomaly;
we use zero-padding at the boundaries}}
\end{tabular}
\end{center}
\end{table*}

\begin{table*}
\begin{center}
\caption{Relaxed confusion matrix for sequential data, with tolerance in predictions}
\label{tblRelConfMatP}
\small
\begin{tabular}{r c c c}
\toprule
                & AA & AnA & \\
\midrule
PA$\delta$ & $\displaystyle\sum_t e_t \mathcal{I}\left(
				\sum_{t'=t-\delta}^{t+\delta}\mathcal{I}(z_{t'} \ge \tau)>0\right)$
           & $\displaystyle\sum_t (1-e_t) \mathcal{I}\left(
				\sum_{t'=t-\delta}^{t+\delta}\mathcal{I}(z_{t'} \ge \tau)>0\right)$
           & $\displaystyle\sum_t \mathcal{I}\left(
				\sum_{t'=t-\delta}^{t+\delta}\mathcal{I}(z_{t'} \ge \tau)>0\right)$ \\
PnA$\delta$& $\displaystyle\sum_t e_t \left(1-\mathcal{I}\left(
				\sum_{t'=t-\delta}^{t+\delta}\mathcal{I}(z_{t'} \ge \tau)>0\right)\right)$
           & $\displaystyle\sum_t (1-e_t) \left(1-\mathcal{I}\left(
				\sum_{t'=t-\delta}^{t+\delta}\mathcal{I}(z_{t'} \ge \tau)>0\right)\right)$
           & $\displaystyle\sum_t \left(1-\mathcal{I}\left(
				\sum_{t'=t-\delta}^{t+\delta}\mathcal{I}(z_{t'} \ge \tau)>0\right)\right)$ \\
\midrule
           & $\displaystyle\sum_t e_t$
           & $\displaystyle\sum_t (1-e_t)$
           & $T$ \\
\bottomrule
\multicolumn{4}{c}{\tiny{%
AA: actual anomaly, AnA: actually no anomaly,
PA$\delta$: predicted anomaly with tolerance $\delta$,
PnA$\delta$: predicted no anomaly with tolerance $\delta$;
we use zero-padding at the boundaries}}
\end{tabular}
\end{center}
\end{table*}

When introducing temporal tolerance in the confusion matrix, we have to make sure that
the result is still a partition with interpretable entries and marginals.
Tables~\ref{tblRelConfMatG} and~\ref{tblRelConfMatP} show the confusion matrices obtained
when introducing temporal tolerance either into the ground-truth time steps or the predicted
time steps, using the formal notation introduced above.
Both confusion matrices partition the observations, but not all entries and marginals
have straightforward interpretations.
Some of the measures usually computed from confusion matrices are therefore uninformative.
The tolerant precision from Equation~\ref{eqnPrecTolerant}
is the $TP$ entry from Table~\ref{tblRelConfMatG} (PA-AA$\delta$) normalized by
the marginal row sum (PA), whereas the tolerant recall from Equation~\ref{eqnRecTolerant}
is given by the $TP$ entry from Table~\ref{tblRelConfMatP} (PA$\delta$-AA) normalized the
marginal column sum (AA).
In both cases, the $TP$ entries and normalization terms are interpretable and yield
informative evaluation measures. We thus restrict our analysis to these two cases
and defer other measures to future work.

\subsection{Statistical significance}

They key question when analyzing evaluation measures from a statistical point of view
is whether the reported values are statistically significant.
To assess statistical significance, we have to treat the quantities in the confusion matrix
as \emph{random variables} that follow some probability distribution.
Only if the reported number of true positives (or any other entry of the confusion matrix) is
\emph{much larger (or smaller) than expected} due to random coincidences, the result should be
considered statistically significant.
\citet{Donges2016} have derived the probability distribution of the two types of true positives
(PA-AA$\delta$ and PA$\delta$-AA) from Tables~\ref{tblRelConfMatG} and~\ref{tblRelConfMatP},
under the assumption that the ground-truth anomalies and the predicted anomalies follow independent
Bernoulli processes. In this case, both quantities follow simple binomial distributions.
\citet{Scharwachter2020a} have generalized the formal analysis for a larger class of problems,
where the anomaly score is a strictly stationary process. They show that in this case PA$\delta$-AA
from Table~\ref{tblRelConfMatP} also follows a binomial distribution, where the success probability
can be approximated using a result from Extreme Value Theory \cite{Coles2001}.
Unfortunately, there is no analogous derivation for PA-AA$\delta$ from Table~\ref{tblRelConfMatG}
under the strict stationarity assumption.
In this work, we do not use the existing analytical results, but perform Monte Carlo simulations
to estimate the required probability distributions without potentially limiting assumptions on
the data generating processes.

 \section{Simulation Study}
\label{secSimulationStudy}

We now use the anomaly score $(z_t)_{t=1,...,T}$ from the earthquake detection example
in Section~\ref{secIntro} and compute time-tolerant confusion matrices, as well as
the time-tolerant precision and recall measures, for randomized ground-truth sequences
of anomalies.
We evaluate the anomaly score against 10,000 random permutations of
the ground-truth sequence of anomalies $(e_t)_{t=1,...,T}$ from the example.
In doing so, we keep the number of ground-truth anomalies constant and assume that they follow
a Bernoulli process. We believe that this assumption is reasonable for ground-truth
anomalies, which typically occur rarely and are not clustered.

\subsection{Monte Carlo precision and recall}

\begin{figure}[tbp]
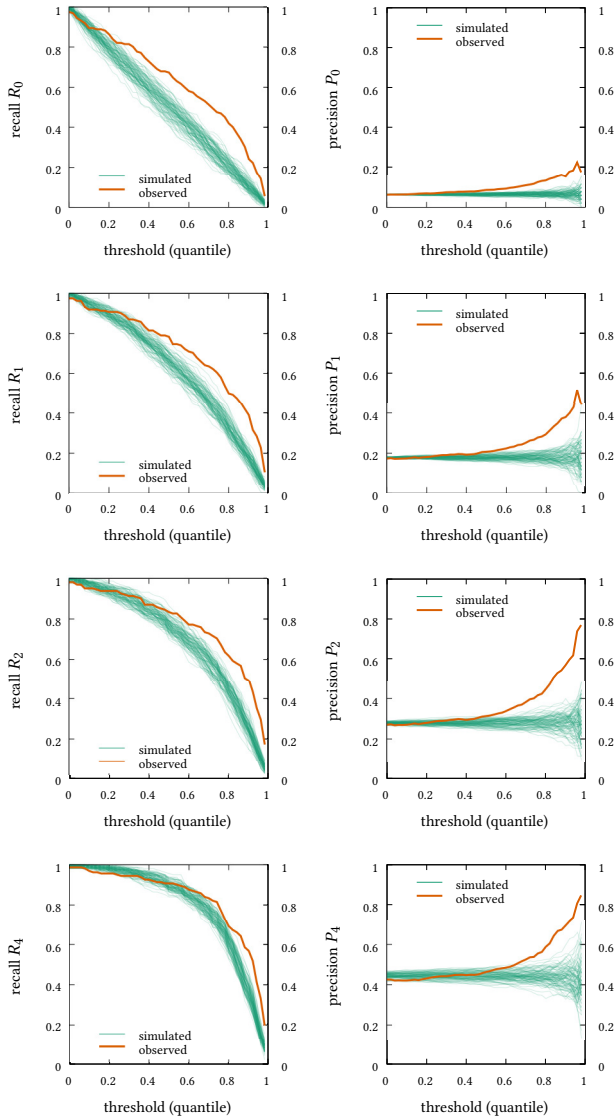

\begin{center}
\begin{gnuplot}[terminal=cairolatex,terminaloptions={size 3.33537,6.0}]
set multiplot layout 4,2
set format y '\tiny{
set format y2 '\tiny{
set format x '\tiny{
set tics scale 0.5
set ytics 0,0.2 offset 0.5,0
set y2tics 0,0.2
set yrange [0:1]
set xtics 0,0.2 offset 0,0.5
set xlabel '\scriptsize{threshold (quantile)}' offset 0,1
#set tmargin screen 0.90
#set bmargin screen 0.20
set lmargin 6
set rmargin 2

set style line 1 lc rgb '#cc1b9e77' lw 1
set style line 2 lc rgb '#00d95f02' lw 3
set style line 3 lc rgb '#001b9e77' lw 1

simuls = 100
array ds = [0,1,2,4]
do for [k=1:|ds|] {
  set ylabel sprintf('\scriptsize{recall $R_
  #$
  set key bottom left samplen 1 Left reverse spacing 0.5
  plot for [i=1:simuls] sprintf("data/earthquakes-simul-recalls-d
       (1/0) w l ls 3 title '\tiny{simulated}', \
       "data/earthquakes-recall.dat" u 1:(column(k+1)) w l ls 2 title '\tiny{observed}'

  set ylabel sprintf('\scriptsize{precision $P_
  #$
  set key top left samplen 1 Left reverse spacing 0.5
  plot for [i=1:simuls] sprintf("data/earthquakes-simul-precisions-d
       (1/0) w l ls 3 title '\tiny{simulated}', \
       "data/earthquakes-precision.dat" u 1:(column(k+1)) w l ls 2 title '\tiny{observed}'
}
\end{gnuplot}
\caption{Simulated and observed values for the precision $P_\delta$ and recall $R_\delta$,
for $\delta \in \{0,1,2,4\}$ and various thresholds.}
\label{figSimulDeltaTau}
\end{center}
\end{figure}

First, we visualize the precision and recall values obtained from
a subset of 100 random permutations for various temporal tolerances and thresholds
in Figure~\ref{figSimulDeltaTau}.
The visualization also shows the performance measures observed on the non-permuted ground-truth
sequence of anomalies.

The observed precision and recall values on the non-permuted sequence are generally higher
than the values from the randomly permuted sequences, especially at larger thresholds.
This confirms that the anomaly score contains useful information on earthquake occurrences.
However, when the temporal tolerance is increased, the gap between the simulated and the
observed performance measures tends to shrink: the performance measures on the simulated
sequences increase to a stronger degree than the performance measures on the observed sequence.
The consequence is that reported performance measures, in particular when computed with
a high temporal tolerance, may not reflect the actual performance of the anomaly detector.
In the worst case, one might conclude that the anomaly score allows detection of anomalies
that are \emph{statistically independent} of the anomaly score.

\subsection{Null distributions}

The simulations clearly show that assessment of the statistical significance of the observed
performance measures is imperative.
For this purpose, we fix the temporal tolerance to $\delta = 2$ and set the threshold $\tau$
to the .9-quantile of the anomaly score.
We observe PA$\delta$-AA${}=80$ (recall $R_\delta=.49$) and
PA-AA$\delta=145$ (precision $P_\delta=.56$)
on the non-permuted ground-truth anomaly sequence from our example.
To assess the statistical significance of the reported numbers, we now have a closer look at the
null distributions for the performance measures obtained in the Monte Carlo simulations.

\begin{figure}[tb]
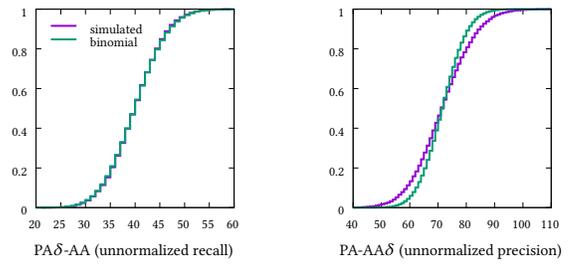

\begin{center}
\begin{gnuplot}[terminal=cairolatex,terminaloptions={size 3.33537,1.5}]
set multiplot layout 1,2
set format y '\tiny{
set format x '\tiny{
set tics scale 0.5
set ytics 0, 0.2 offset 0.5,0
set xtics offset 0,0.5
set tmargin screen 0.90
set bmargin screen 0.20
set lmargin 5
set rmargin 3

set xlabel '\scriptsize{PA$\delta$-AA (unnormalized recall)}' offset 0,1
set xrange [20:60]
set key samplen 1 spacing 0.5 Left reverse at 0, 0.95
plot "data/earthquakes-recall-distrib-d2.dat" u 1:2 w l lw 3 title '\tiny{simulated}', \
     "data/earthquakes-recall-distrib-d2.dat" u 3:4 w l lw 3 title '\tiny{binomial}'

set xlabel '\scriptsize{PA-AA$\delta$ (unnormalized precision)}' offset 0,1
set xrange [40:110]
plot "data/earthquakes-precision-distrib-d2.dat" u 1:2 w l lw 3 title '\tiny{simulated}', \
     "data/earthquakes-precision-distrib-d2.dat" u 3:4 w l lw 3 title '\tiny{binomial}'
\end{gnuplot}
\caption{Cumulative distribution functions of the two types of true positives required
for precision and recall.}
\label{figNullDistr}
\end{center}
\end{figure}

Figure~\ref{figNullDistr} (simulated) shows the cumulative distribution functions
for the numbers of true positives obtained from 10,000 simulations for the specific choice of
$\delta$ and $\tau$ mentioned above.
Given the simulated distributions, we can easily compute Monte Carlo $p$-values
\citep{Davison1997} for the numbers of true positives:
The $p$-value is the probability that we obtain a value for
the true positive at least as high as the observed one.
Since the performance measures were smaller than the reported values in \emph{all} of the simulated
runs, we have $p < .0001$ for both precision and recall, which is highly significant.

The analytical null distributions derived in the literature \cite{Donges2016,Scharwachter2020a}
are all binomial. To complete our analysis, we now check whether our simulations also yield
binomial distributions.
Figure~\ref{figNullDistr} (binomial) shows the cumulative distribution functions of binomial
random variables when the binomial success probabilities are estimated from our Monte Carlo
simulations.
The plots suggest that the true positive PA$\delta$-AA for the recall follows a binomial
distribution, whereas the true positive PA-AA$\delta$ for the precision seems to be overdispersed
with respect to the binomial distribution (it has a larger variance). We have repeated
the experiment with different thresholds and temporal tolerances and observed the same behavior
across all experiments. The exact form of the overdispersed distribution should be investigated
more deeply in future work.

\section{Conclusion}

We have presented time-tolerant variants of the precision and recall measures routinely used
to evaluate anomaly detectors for sequential data.
We have shown that these measures are computed from two distinct time-tolerant confusion matrices.
Time-tolerant confusion matrices can, in principle, be used to derive time-tolerant
variants of other well-known measures. However, care has to be taken to preserve interpretability.
We applied the time-tolerant precision and recall measures on an example anomaly detection
problem, and analyzed their statistical behaviors in a simulation study.
Our experiments suggest that reported values for precision and recall can overestimate the
performance of an anomaly detector even with moderate temporal tolerances.
We have demonstrated how to obtain Monte Carlo $p$-values to assess the statistical significance
of reported performance measures, using randomly permuted ground-truth sequences.
We believe that establishing the statistical significance of reported precision and recall
values should become a community standard. Future work should improve the analytical
understanding of the null distributions required for this task.

\balance

\bibliographystyle{ACM-Reference-Format}
\bibliography{ms}


\begin{thebibliography}{23}


\ifx \showCODEN    \undefined \def \showCODEN     #1{\unskip}     \fi
\ifx \showDOI      \undefined \def \showDOI       #1{#1}\fi
\ifx \showISBNx    \undefined \def \showISBNx     #1{\unskip}     \fi
\ifx \showISBNxiii \undefined \def \showISBNxiii  #1{\unskip}     \fi
\ifx \showISSN     \undefined \def \showISSN      #1{\unskip}     \fi
\ifx \showLCCN     \undefined \def \showLCCN      #1{\unskip}     \fi
\ifx \shownote     \undefined \def \shownote      #1{#1}          \fi
\ifx \showarticletitle \undefined \def \showarticletitle #1{#1}   \fi
\ifx \showURL      \undefined \def \showURL       {\relax}        \fi
\providecommand\bibfield[2]{#2}
\providecommand\bibinfo[2]{#2}
\providecommand\natexlab[1]{#1}
\providecommand\showeprint[2][]{arXiv:#2}

\bibitem[\protect\citeauthoryear{Adams and Marlin}{Adams and Marlin}{2017}]%
        {Adams2017}
\bibfield{author}{\bibinfo{person}{Roy~J. Adams} {and}
  \bibinfo{person}{Benjamin~M. Marlin}.} \bibinfo{year}{2017}\natexlab{}.
\newblock \showarticletitle{{Learning time series detection models from
  temporally imprecise labels}}. In \bibinfo{booktitle}{\emph{AISTATS}}.
\newblock


\bibitem[\protect\citeauthoryear{Adams and Marlin}{Adams and Marlin}{2018}]%
        {Adams2018a}
\bibfield{author}{\bibinfo{person}{Roy~J. Adams} {and}
  \bibinfo{person}{Benjamin~M. Marlin}.} \bibinfo{year}{2018}\natexlab{}.
\newblock \showarticletitle{{Learning time series segmentation models from
  temporally imprecise labels}}. In \bibinfo{booktitle}{\emph{UAI}}.
\newblock


\bibitem[\protect\citeauthoryear{Bridges, Jamieson, and Reed}{Bridges
  et~al\mbox{.}}{2017}]%
        {Bridges2017}
\bibfield{author}{\bibinfo{person}{Robert~A Bridges}, \bibinfo{person}{Jessie~D
  Jamieson}, {and} \bibinfo{person}{Joel~W Reed}.}
  \bibinfo{year}{2017}\natexlab{}.
\newblock \showarticletitle{{Setting the threshold for high throughput
  detectors}}. In \bibinfo{booktitle}{\emph{IEEE Big Data}}.
\newblock


\bibitem[\protect\citeauthoryear{Chi, Han, and Wang}{Chi et~al\mbox{.}}{2016}]%
        {Chi2016a}
\bibfield{author}{\bibinfo{person}{Lianhua Chi}, \bibinfo{person}{Bo Han},
  {and} \bibinfo{person}{Yun Wang}.} \bibinfo{year}{2016}\natexlab{}.
\newblock \showarticletitle{{Open Problem: Accurately Measuring Event Impacts
  on Time Series}}. In \bibinfo{booktitle}{\emph{KDD MiLeTS Workshop}}.
\newblock


\bibitem[\protect\citeauthoryear{Coles}{Coles}{2001}]%
        {Coles2001}
\bibfield{author}{\bibinfo{person}{Stuart Coles}.}
  \bibinfo{year}{2001}\natexlab{}.
\newblock \bibinfo{booktitle}{\emph{{An Introduction to Statistical Modeling of
  Extreme Values}}}.
\newblock \bibinfo{publisher}{Springer-Verlag London, Ltd},
  \bibinfo{address}{London, UK}.
\newblock


\bibitem[\protect\citeauthoryear{Davison and Hinkley}{Davison and
  Hinkley}{1997}]%
        {Davison1997}
\bibfield{author}{\bibinfo{person}{A.~C. Davison} {and} \bibinfo{person}{D.~V.
  Hinkley}.} \bibinfo{year}{1997}\natexlab{}.
\newblock \bibinfo{booktitle}{\emph{{Bootstrap Methods and Their
  Application}}}.
\newblock \bibinfo{publisher}{Cambridge University Press},
  \bibinfo{address}{Cambridge, UK}.
\newblock


\bibitem[\protect\citeauthoryear{Donges, Schleusner, Siegmund, and
  Donner}{Donges et~al\mbox{.}}{2016}]%
        {Donges2016}
\bibfield{author}{\bibinfo{person}{J.~F. Donges}, \bibinfo{person}{C.-F.
  Schleusner}, \bibinfo{person}{J.~F. Siegmund}, {and} \bibinfo{person}{R.~V.
  Donner}.} \bibinfo{year}{2016}\natexlab{}.
\newblock \showarticletitle{{Event coincidence analysis for quantifying
  statistical interrelationships between event time series: On the role of
  flood events as triggers of epidemic outbreaks}}.
\newblock \bibinfo{journal}{\emph{European Physics Journal Special Topics}}
  \bibinfo{volume}{487} (\bibinfo{year}{2016}), \bibinfo{pages}{471--487}.
\newblock


\bibitem[\protect\citeauthoryear{Earle, Bowden, and Guy}{Earle
  et~al\mbox{.}}{2011}]%
        {Earle2011}
\bibfield{author}{\bibinfo{person}{Paul~S. Earle}, \bibinfo{person}{Daniel~C.
  Bowden}, {and} \bibinfo{person}{Michelle Guy}.}
  \bibinfo{year}{2011}\natexlab{}.
\newblock \showarticletitle{{Twitter earthquake detection: Earthquake
  monitoring in a social world}}.
\newblock \bibinfo{journal}{\emph{Annals of Geophysics}} \bibinfo{volume}{54},
  \bibinfo{number}{6} (\bibinfo{year}{2011}), \bibinfo{pages}{708--715}.
\newblock


\bibitem[\protect\citeauthoryear{Luo, Lou, Lin, Fu, Ding, Zhang, and Wang}{Luo
  et~al\mbox{.}}{2014}]%
        {Luo2014}
\bibfield{author}{\bibinfo{person}{Chen Luo}, \bibinfo{person}{Jian-Guang Lou},
  \bibinfo{person}{Qingwei Lin}, \bibinfo{person}{Qiang Fu},
  \bibinfo{person}{Rui Ding}, \bibinfo{person}{Dongmei Zhang}, {and}
  \bibinfo{person}{Zhe Wang}.} \bibinfo{year}{2014}\natexlab{}.
\newblock \showarticletitle{{Correlating events with time series for incident
  diagnosis}}. In \bibinfo{booktitle}{\emph{KDD}}.
\newblock


\bibitem[\protect\citeauthoryear{Malhotra, Vig, Shroff, and Agarwal}{Malhotra
  et~al\mbox{.}}{2015}]%
        {Malhotra2015}
\bibfield{author}{\bibinfo{person}{Pankaj Malhotra}, \bibinfo{person}{Lovekesh
  Vig}, \bibinfo{person}{Gautam Shroff}, {and} \bibinfo{person}{Puneet
  Agarwal}.} \bibinfo{year}{2015}\natexlab{}.
\newblock \showarticletitle{{Long Short Term Memory Networks for Anomaly
  Detection in Time Series}}. In \bibinfo{booktitle}{\emph{ESANN}}.
\newblock


\bibitem[\protect\citeauthoryear{Powers}{Powers}{2007}]%
        {Powers2007}
\bibfield{author}{\bibinfo{person}{David M.~W. Powers}.}
  \bibinfo{year}{2007}\natexlab{}.
\newblock \bibinfo{booktitle}{\emph{{Evaluation: From Precision, Recall and
  F-Factor to ROC, Informedness, Markedness {\&} Correlation}}}.
\newblock \bibinfo{type}{{T}echnical {R}eport}. \bibinfo{institution}{Flinders
  University of South Australia}, \bibinfo{address}{Adelaide, Australia}.
\newblock


\bibitem[\protect\citeauthoryear{Quiroga, Kreuz, and Grassberger}{Quiroga
  et~al\mbox{.}}{2002}]%
        {Quiroga2002}
\bibfield{author}{\bibinfo{person}{R.~Quian Quiroga}, \bibinfo{person}{T.
  Kreuz}, {and} \bibinfo{person}{P. Grassberger}.}
  \bibinfo{year}{2002}\natexlab{}.
\newblock \showarticletitle{{Event synchronization: A simple and fast method to
  measure synchronicity and time delay patterns}}.
\newblock \bibinfo{journal}{\emph{Physical Review E}} \bibinfo{volume}{66},
  \bibinfo{number}{4} (\bibinfo{year}{2002}).
\newblock


\bibitem[\protect\citeauthoryear{Ren, Xu, Wang, Yi, Huang, Kou, Xing, Yang,
  Tong, and Zhang}{Ren et~al\mbox{.}}{2019}]%
        {Ren2019}
\bibfield{author}{\bibinfo{person}{Hansheng Ren}, \bibinfo{person}{Bixiong Xu},
  \bibinfo{person}{Yujing Wang}, \bibinfo{person}{Chao Yi},
  \bibinfo{person}{Congrui Huang}, \bibinfo{person}{Xiaoyu Kou},
  \bibinfo{person}{Tony Xing}, \bibinfo{person}{Mao Yang}, \bibinfo{person}{Jie
  Tong}, {and} \bibinfo{person}{Qi Zhang}.} \bibinfo{year}{2019}\natexlab{}.
\newblock \showarticletitle{{Time-Series Anomaly Detection Service at
  Microsoft}}. In \bibinfo{booktitle}{\emph{KDD}}.
\newblock


\bibitem[\protect\citeauthoryear{Scharw{\"{a}}chter and
  M{\"{u}}ller}{Scharw{\"{a}}chter and M{\"{u}}ller}{2020a}]%
        {Scharwachter2020a}
\bibfield{author}{\bibinfo{person}{Erik Scharw{\"{a}}chter} {and}
  \bibinfo{person}{Emmanuel M{\"{u}}ller}.} \bibinfo{year}{2020}\natexlab{a}.
\newblock \showarticletitle{{Does Terrorism Trigger Online Hate Speech? On the
  Association of Events and Time Series}}.
\newblock \bibinfo{journal}{\emph{Annals of Applied Statistics}}
  (\bibinfo{year}{2020}).
\newblock


\bibitem[\protect\citeauthoryear{Scharw{\"{a}}chter and
  M{\"{u}}ller}{Scharw{\"{a}}chter and M{\"{u}}ller}{2020b}]%
        {Scharwachter2020}
\bibfield{author}{\bibinfo{person}{Erik Scharw{\"{a}}chter} {and}
  \bibinfo{person}{Emmanuel M{\"{u}}ller}.} \bibinfo{year}{2020}\natexlab{b}.
\newblock \showarticletitle{{Two-Sample Testing for Event Impacts in Time
  Series}}. In \bibinfo{booktitle}{\emph{Proceedings of the SIAM International
  Conference on Data Mining (SIAM SDM)}}.
\newblock


\bibitem[\protect\citeauthoryear{Siegmund, Wiedermann, Donges, and
  Donner}{Siegmund et~al\mbox{.}}{2016}]%
        {Siegmund2016}
\bibfield{author}{\bibinfo{person}{Jonatan~F. Siegmund}, \bibinfo{person}{Marc
  Wiedermann}, \bibinfo{person}{Jonathan~F. Donges}, {and}
  \bibinfo{person}{Reik~V. Donner}.} \bibinfo{year}{2016}\natexlab{}.
\newblock \showarticletitle{{Impact of temperature and precipitation extremes
  on the flowering dates of four German wildlife shrub species}}.
\newblock \bibinfo{journal}{\emph{Biogeosciences}} (\bibinfo{year}{2016}).
\newblock


\bibitem[\protect\citeauthoryear{Siffer, Fouque, Termier, and Largouet}{Siffer
  et~al\mbox{.}}{2017}]%
        {Siffer2017}
\bibfield{author}{\bibinfo{person}{A. Siffer}, \bibinfo{person}{P.-A. Fouque},
  \bibinfo{person}{A. Termier}, {and} \bibinfo{person}{C. Largouet}.}
  \bibinfo{year}{2017}\natexlab{}.
\newblock \showarticletitle{{Anomaly detection in streams with extreme value
  theory}}. In \bibinfo{booktitle}{\emph{KDD}}.
\newblock


\bibitem[\protect\citeauthoryear{Su, Liu, Zhao, Sun, Niu, and Pei}{Su
  et~al\mbox{.}}{2019}]%
        {Su2019}
\bibfield{author}{\bibinfo{person}{Ya Su}, \bibinfo{person}{Rong Liu},
  \bibinfo{person}{Youjian Zhao}, \bibinfo{person}{Wei Sun},
  \bibinfo{person}{Chenhao Niu}, {and} \bibinfo{person}{Dan Pei}.}
  \bibinfo{year}{2019}\natexlab{}.
\newblock \showarticletitle{{Robust anomaly detection for multivariate time
  series through stochastic recurrent neural network}}. In
  \bibinfo{booktitle}{\emph{KDD}}.
\newblock


\bibitem[\protect\citeauthoryear{Tatbul, Lee, Zdonik, Alam, and
  Gottschlich}{Tatbul et~al\mbox{.}}{2018}]%
        {Tatbul2018}
\bibfield{author}{\bibinfo{person}{Nesime Tatbul}, \bibinfo{person}{Tae~Jun
  Lee}, \bibinfo{person}{Stan Zdonik}, \bibinfo{person}{Mejbah Alam}, {and}
  \bibinfo{person}{Justin Gottschlich}.} \bibinfo{year}{2018}\natexlab{}.
\newblock \showarticletitle{{Precision and recall for time series}}. In
  \bibinfo{booktitle}{\emph{NeurIPS}}.
\newblock
\showISSN{10495258}


\bibitem[\protect\citeauthoryear{van Dortmont, van~den Elzen, and van Wijk}{van
  Dortmont et~al\mbox{.}}{2019}]%
        {VanDortmont2019}
\bibfield{author}{\bibinfo{person}{M.~A.M.M. van Dortmont}, \bibinfo{person}{S.
  van~den Elzen}, {and} \bibinfo{person}{J.~J. van Wijk}.}
  \bibinfo{year}{2019}\natexlab{}.
\newblock \showarticletitle{{ChronoCorrelator: Enriching events with time
  series}}. In \bibinfo{booktitle}{\emph{EuroVis}}.
\newblock


\bibitem[\protect\citeauthoryear{Wiedermann, Radebach, Donges, Kurths, and
  Donner}{Wiedermann et~al\mbox{.}}{2016}]%
        {Wiedermann2016}
\bibfield{author}{\bibinfo{person}{Marc Wiedermann}, \bibinfo{person}{Alexander
  Radebach}, \bibinfo{person}{Jonathan~F. Donges},
  \bibinfo{person}{J{\"{u}}rgen Kurths}, {and} \bibinfo{person}{Reik~V.
  Donner}.} \bibinfo{year}{2016}\natexlab{}.
\newblock \showarticletitle{{A climate network-based index to discriminate
  different types of El Ni{\~{n}}o and La Ni{\~{n}}a}}.
\newblock \bibinfo{journal}{\emph{Geophysical Research Letters}}
  \bibinfo{volume}{43}, \bibinfo{number}{13} (\bibinfo{year}{2016}),
  \bibinfo{pages}{7176--7185}.
\newblock


\bibitem[\protect\citeauthoryear{Withers, Aster, Young, Beiriger, Harris,
  Moore, and Trujillo}{Withers et~al\mbox{.}}{1998}]%
        {Withers1998}
\bibfield{author}{\bibinfo{person}{Mitchell Withers}, \bibinfo{person}{Richard
  Aster}, \bibinfo{person}{Christopher Young}, \bibinfo{person}{Judy Beiriger},
  \bibinfo{person}{Mark Harris}, \bibinfo{person}{Susan Moore}, {and}
  \bibinfo{person}{Julian Trujillo}.} \bibinfo{year}{1998}\natexlab{}.
\newblock \showarticletitle{{A comparison of select trigger algorithms for
  automated global seismic phase and event detection}}.
\newblock \bibinfo{journal}{\emph{Bulletin of the Seismological Society of
  America}} \bibinfo{volume}{88}, \bibinfo{number}{1} (\bibinfo{year}{1998}),
  \bibinfo{pages}{95--106}.
\newblock
\showISSN{00371106}


\bibitem[\protect\citeauthoryear{Xu, Feng, Chen, Wang, Qiao, Chen, Zhao, Li,
  Bu, Li, Liu, Zhao, and Pei}{Xu et~al\mbox{.}}{2018}]%
        {Xu2018a}
\bibfield{author}{\bibinfo{person}{Haowen Xu}, \bibinfo{person}{Yang Feng},
  \bibinfo{person}{Jie Chen}, \bibinfo{person}{Zhaogang Wang},
  \bibinfo{person}{Honglin Qiao}, \bibinfo{person}{Wenxiao Chen},
  \bibinfo{person}{Nengwen Zhao}, \bibinfo{person}{Zeyan Li},
  \bibinfo{person}{Jiahao Bu}, \bibinfo{person}{Zhihan Li},
  \bibinfo{person}{Ying Liu}, \bibinfo{person}{Youjian Zhao}, {and}
  \bibinfo{person}{Dan Pei}.} \bibinfo{year}{2018}\natexlab{}.
\newblock \showarticletitle{{Unsupervised Anomaly Detection via Variational
  Auto-Encoder for Seasonal KPIs in Web Applications}}. In
  \bibinfo{booktitle}{\emph{WWW}}.
\newblock


\end{thebibliography}

\end{document}